\documentclass[runningheads]{llncs}
\makeatletter
\newcommand{\printfnsymbol}[1]{%
  \textsuperscript{\@fnsymbol{#1}}%
}
\makeatother
\usepackage{graphicx}
\usepackage{multirow}
\usepackage{amssymb}
\usepackage{amsmath}
\usepackage{enumerate}
\usepackage[inline]{enumitem}
\usepackage{hyperref}
\usepackage[separate-uncertainty,bracket-numbers = false, table-align-uncertainty = true, table-align-exponent=false]{siunitx}
\usepackage{multirow}
\usepackage{adjustbox}
\DeclareMathOperator*{\argmin}{argmin}

\begin{document}
\title{Temporal coherence-based self-supervised learning for laparoscopic workflow analysis}
\titlerunning{Temporal coherence-based self-supervised learning for lap. workflow analysis}
%
\author{Isabel Funke\inst{1}\thanks{Both authors contributed equally to this work.} \and Alexander Jenke\inst{1} \and S{\"o}ren Torge Mees\inst{2} \and J{\"urgen} Weitz\inst{2} \and Stefanie Speidel\inst{1} \and Sebastian Bodenstedt\inst{1}\printfnsymbol{1}}
\authorrunning{I. Funke et al.}
%
\institute{Department for Translational Surgical Oncology, National Center for Tumor Diseases (NCT), Partner Site Dresden, Dresden, Germany\\ \email{Firstname.Lastname@nct-dresden.de}\\ \and Department of Visceral, Thoracic and Vascular Surgery, Faculty of Medicine and University Hospital Carl Gustav Carus, TU Dresden, Dresden, Germany\\ }
\maketitle              
\begin{abstract}
In order to provide the right type of assistance at the right time, computer-assisted surgery systems need context awareness.
To \linebreak achieve this, methods for surgical workflow analysis are crucial. 
Currently, convolutional neural networks provide the best performance for video-based workflow analysis tasks.
For training such networks, large amounts of annotated data are necessary.
However, collecting a sufficient amount of data is often costly,  time-consuming, and not always feasible.
In this paper, we address this problem by presenting and comparing different approaches for self-supervised pretraining of neural networks on unlabeled laparoscopic videos using temporal coherence.
We evaluate our pretrained networks on Cholec80, a publicly available dataset for surgical phase segmentation, on which a maximum $F_1$ score of 84.6 was reached.
Furthermore, we were able to achieve an increase of the $F_1$ score of up to 10 points when compared to a non-pretrained neural network.


\end{abstract}
\section{Introduction}
The aim of a computer-assisted surgery (CAS) system is to provide the surgeon with the right type of assistance at the right time. 
To achieve this, context awareness is crucial. 
This means that the system must be able to understand the processes currently taking place in the operating room (OR) and adapt its behavior accordingly.
\emph{Surgical workflow analysis} covers the challenging topic of perceiving, understanding, and describing surgical processes~\cite{lalys2014surgical}.

A common approach is to analyze surgical processes by interpreting a time series of signals that are recorded by sensors -- in some cases also by humans -- in the OR.
As laparoscopies are performed via camera, methods that require only video as input sensor data are of special interest, since the video can be collected effortlessly during surgery.

State-of-the-art video-based approaches for workflow analysis rely on deep neural networks~\cite{aksamentov2017deep,Bodenstedt2017,jin2018sv,TwinandaSMMMP16,yengera2018less}.
However, deep learning-based methods require large amounts of labeled data for training. Especially in surgery, obtaining a sufficient amount of annotated video data is difficult and costly.

To alleviate the problem of limited training data, it is common to \emph{pretrain} neural networks and fine-tune them afterwards.
Often, networks are pretrained using labeled data coming from another domain, such as ImageNet~\cite{deng2009imagenet}.
Another way is to use unlabeled data from the same domain and train on a proxy task using labels inherent in the data, which is called \emph{self-supervised} learning.

For self-supervised learning from video, a number of ideas have been proposed~\cite{goroshin2015unsupervised,jayaraman2016slow,lee2017unsupervised,misra2016shuffle,wang2015unsupervised}.
Most exploit the \emph{temporal coherence} of video, which implies that
\begin{enumerate*}[label=(\roman*)]
\item consecutive frames are in temporal order,
\item frames change slowly over time, and
\item frames change steadily, i.e., abrupt motions are unlikely.
\end{enumerate*}   

The studies~\cite{lee2017unsupervised} and~\cite{misra2016shuffle} propose proxy tasks based on the temporal order between frames.
In line with this, \cite{Bodenstedt2017} use the task to order pairs of laparoscopic images for pretraining a network for surgical phase segmentation.
\emph{Surgical phase segmentation}~\cite{Padoy2012632} is the problem of recognizing the surgical phase being performed by the surgeon at each point during surgery.
Another proxy task for this problem is to predict the progress and remaining duration of a surgery~\cite{yengera2018less}. 

Intuitively, these tasks encourage the network to learn discriminative features that are useful to infer the absolute or relative temporal position of a video frame.
In contrast, \cite{goroshin2015unsupervised,jayaraman2016slow,wang2015unsupervised} aim at learning features that are \emph{invariant} to typical alterations occurring between adjacent frames, such as slight rotations or deformations. To this end, they aim to ensure that temporally close frames, which most likely depict the same semantic scene, are mapped to similar representations in feature space. This idea goes back to \emph{Slow Feature Analysis~(SFA)}~\cite{wiskott2002slow}.

In this paper, we describe and compare different approaches to exploit temporal coherence while pretraining a convolutional neural network (CNN) for surgical phase segmentation. 
We assume the pretraining encourages the CNN to learn features that are invariant to irrelevant changes between adjacent frames, such as slight movements of instruments or of the endoscope, while being discriminative enough to distinguish between semantically different frames.

To promote reproducibility and to fuel future research, we made our code available at \url{https://gitlab.com/nct_tso_public/pretrain_tc}.

Experiments using the Cholec80 dataset \cite{TwinandaSMMMP16} demonstrate that a CNN pretrained to exploit the temporal coherence of unlabeled laparoscopic video outperforms a non-pretrained CNN after being fine-tuned for surgical phase segmentation.
When only 20 labeled videos are available, the proposed pretraining achieves an increase from 67.8 to 78.6 as measured by $F_1$ score.

\section{Methods}
\label{methods}
The core of our neural network architecture for surgical phase segmentation is a ResNet-50 CNN \cite{he2016deep}. We initialize it with ImageNet~\cite{deng2009imagenet} pretrained weights and further train it on unlabeled videos of laparoscopic surgeries, using an SFA-based approach for self-supervised learning.
This encourages the CNN to map temporally close video frames to similar representations in feature space. 

More formally, the CNN learns an embedding $f:\mathbb{R}^{3 \times h \times w}\rightarrow \mathbb{R}^{d}$, where $\mathbb{R}^{d}$ is the $d$-dimensional feature space and $\mathbb{R}^{3 \times h \times w}$ is the space of laparoscopic video frames with height~$h$, width~$w$, and three color channels (RGB). 
Let $I_t \in \mathbb{R}^{3 \times h \times w}$ denote the frame at time step $t$. 
To suffice temporal coherence, we require that $f(I_t) \approx f(I_{t + \Delta})$ for a small $\Delta$ with $|\Delta| < \delta$.
To learn an embedding that is discriminative and to avoid trivial solutions such as $f(I_t):= 0$, we require that $f(I_t)$ and $f(I_{t + \Gamma})$ lie further apart in feature space when $\Gamma$ is large, i.e., $|\Gamma| > \gamma$ (see subsection~\ref{methods:pretrain} for details). $\delta$ and $\gamma$ are nonnegative real-valued parameters.

To evaluate the efficacy of the proposed self-supervised pretraining approach, we extend the CNN into a recurrent neural network (RNN) and fine-tune the CNN-RNN for surgical phase segmentation using annotated laparoscopic videos (see subsection~\ref{methods:finetune}). We can then compare the performance of the pretrained CNN-RNN to the performance of a CNN-RNN that has been trained solely for the surgical phase segmentation task (see section~\ref{eval}).

\subsection{Self-supervised pretraining}
\label{methods:pretrain}
For self-supervised pretraining, the output layer of the ResNet-50 CNN is replaced with a  fully connected layer with $d = 4096$ output neurons (\textit{FeatureNet}). 
As the CNN has been pretrained on ImageNet, we only adjust the weights of the \textit{conv5\_x} layers and of the newly added fully connected layer during training. 

Given a frame $I_t$, we calculate the embedding $F_t := f(I_t)$ by forwarding the frame through \emph{FeatureNet} and taking the output $(o_1, o_2, ..., o_d)^T \in \mathbb{R}^{d}$ at the last layer.
We train  \textit{FeatureNet} to learn a temporally coherent video frame embedding using one of the following methods. Throughout this section, $D$ denotes a distance function, in our case the L2 norm.
\begin{enumerate}[label=(\alph*)]
\item Training with \textbf{contrastive} loss \label{method:contrastive}\\ 
Given a video with $T$ frames, we create a tuple $(I_t, I_{t + \Delta}, I_{t + \Gamma})$ by sampling $t$ from $[0, T - 1]$, $\Delta$ from $[-\delta, \delta]$, and $\Gamma$ from $[-(T - 1), -\gamma] \cup [\gamma, T - 1]$ uniformly at random. 
Regarding \emph{FeatureNet} as a Siamese network~\cite{bromley1994signature}, we propagate the temporally close pair $(I_t, I_{t + \Delta})$ through the CNN and calculate $D(F_t, F_{t + \Delta})$. 
Likewise, we propagate the temporally distant pair $(I_t, I_{t + \Gamma})$ and calculate $D(F_t, F_{t + \Gamma})$.  
Finally, we calculate the contrastive loss~\cite{goroshin2015unsupervised} 
$$L_c(F_t, F_{t + \Delta}, F_{t + \Gamma}) = D(F_t, F_{t + \Delta}) + \mathtt{max}\lbrace 0, m_c - D(F_t, F_{t + \Gamma})\rbrace.$$  
This loss function encourages $F_t$ to be close to $F_{t + \Delta}$, while $F_t$ and $F_{t + \Gamma}$ are enforced to be separated by margin $m_c$.
\item Training with \textbf{ranking} loss \\
A training tuple $(I_t, I_{t + \Delta}, I_{t + \Gamma})$ is created the same way as in method~\ref{method:contrastive}. Regarding \emph{FeatureNet} as a Triplet Siamese Network, we propagate the triplet $(I_t, I_{t + \Delta}, I_{t + \Gamma})$ through the CNN and calculate the ranking loss~\cite{wang2015unsupervised}
$$L_r(F_t, F_{t + \Delta}, F_{t + \Gamma}) = \mathtt{max}\lbrace 0, D(F_t, F_{t + \Delta}) - D(F_t, F_{t + \Gamma}) + m_r\rbrace.$$
This loss function considers the distance between $F_t$ and $F_{t + \Delta}$ relative to the distance between $F_t$ and $F_{t + \Gamma}$ and encourages $F_t$ and $F_{t + \Delta}$  to be closer together than $F_t$ and $F_{t + \Gamma}$ by a margin of $m_r$.  
\item Training with \textbf{1\textsuperscript{st} \& 2\textsuperscript{nd} order contrastive} loss \\
While (first order) temporal coherence requires the first order temporal derivatives in the learned feature space to be small, i.e.,  $F_t \approx F_{t + \Delta}$, \emph{second order temporal coherence}~\cite{jayaraman2016slow} requires the second order temporal derivatives to be small, i.e., $F_t - F_{t + \Delta} \approx F_{t + \Delta} - F_{t + 2\Delta}$ for a small value of $\Delta$. \\
Intuitively, first order temporal coherence ensures that embeddings do not change quickly over time, while second order temporal coherence ensures that the changes are consistent, or steady, across neighboring frames.  
Applying the contrastive loss function to second order temporal coherence yields
$$L_{c_2}(F_t, F_{t + \Delta}, F_{t + 2\Delta}, F_{t + \Gamma}) = L_c(F_t - F_{t + \Delta}, F_{t + \Delta} - F_{t + 2\Delta},F_{t + \Delta} - F_{t + \Gamma})$$
In practice, we create a training tuple $(I_t, I_{t + \Delta}, I_{t + 2\Delta}, I_{t + \Gamma})$ by sampling $t$, $\Delta$, and $\Gamma$ as described in method~\ref{method:contrastive}. 
Regarding \emph{FeatureNet} as a Triplet Sia-mese Network, we propagate the triplets $(I_t, I_{t + \Delta}, I_{t + 2\Delta})$ and  $(I_t, I_{t + \Delta}, I_{t + \Gamma})$ through the network and calculate $L_{c_2}$. We then combine it with the first order contrastive loss $L_c$ into an overall loss $L_{c + c_2} = L_c + \omega L_{c_2},$
where $\omega = 0.5$ is a nonnegative real-valued weight parameter.
\end{enumerate}
\subsection{Supervised fine-tuning for surgical phase segmentation}
\label{methods:finetune}
Once pretrained, we modify the CNN for surgical phase segmentation by extending it into an RNN using a long short-term memory unit (LSTM) \cite{hochreiter1997long} with 512 neurons.
The LSTM is followed by a fully connected layer, which has one output neuron per surgical phase.
We refer to this CNN-LSTM as \textit{PhaseNet}.
During fine-tuning, the weights of the CNN and the LSTM are jointly optimized. However, the weights of the ResNet-50 layers below \textit{conv5\_x} stay frozen.

\section{Evaluation}
\label{eval}
For evaluation, we used the publicly available Cholec80 dataset \cite{TwinandaSMMMP16}. It consists of 80 videos from laparoscopic cholecystectomies, annotated with surgical phase labels.
We divided the dataset into four sets A, B, C, and D of equal size and similar average procedure length.
A, B, and C were used for training, while D was withheld for testing. For pretraining, we extracted video frames at 5 Hz. Training and testing for phase segmentation was performed at 1 Hz.
Each frame was downsized to $384 \times 216$ px.

We trained three different versions of \textit{FeatureNet}, one with each of the pretraining variants described in section \ref{methods:pretrain}.
The union of sets A, B, and C (i. e., 60 videos in total) was used as training data, ignoring the labels.
Each CNN was trained for 25 epochs. Per epoch, we randomly sampled 250 tuples per video, which were processed in batches of size~64.
$\delta$~was set to 30 sec (15 sec for variant (c)), $\gamma$ to 120 sec and $m_c = m_r = 2$.
We used the Adam optimizer \cite{kingma2014adam} with a learning rate of $10^{-4}$ .
All newly added layers were initialized with random values from the range $(\frac{-1}{\sqrt n} , \frac{1}{\sqrt n})$, with $n$ being the number of neurons in the layer.

To evaluate the suitability of the proposed pretraining approach for surgical phase segmentation, each of the pretrained CNNs  (\textbf{contrastive}, \textbf{ranking}, and \textbf{1\textsuperscript{st} \& 2\textsuperscript{nd} order contrastive}) was extended into a \textit{PhaseNet} and fine-tuned using the labeled videos from either set A (\#OPs = \textbf{20}), sets A and B (\#OPs = \textbf{40}), or sets A, B, and C (\#OPs = \textbf{60}).
As baseline, a \textit{PhaseNet} without self-supervised pretraining (\textbf{no pretraining}) was fine-tuned in the same manner.
Note that the underlying ResNet-50 CNN had still been pretrained on ImageNet.

For fine-tuning the networks, we used the Adam optimizer \cite{kingma2014adam} with a learning rate of $10^{-4}$ and a batch size of 128.
After every batch, the content of the LSTM's hidden state was saved and restored for the next batch.
Due to hardware restraints, gradients were only accumulated for three batches before applying the optimizer.
Training was stopped once the accuracy on the training set climbed above 99.9\%.
All newly added layers were initialized as described above.

The results of evaluating each \textit{PhaseNet} on test set D can be found in table~\ref{tab:comparison:methods}. We calculated the metrics \emph{accuracy}, \emph{recall}, and \emph{precision} as defined in~\cite{Padoy2012632}. The $F_1$ score is the harmonic mean of precision and recall. The metrics were averaged over all operations in the test set.
Table~\ref{tab:comparison:phases} presents the phase-wise results of the best performing pretrained \textit{PhaseNet} (\textbf{1\textsuperscript{st} \& 2\textsuperscript{nd} order contrastive}) compared to the \emph{PhaseNet} that did not undergo self-supervised pretraining.

\setlength{\tabcolsep}{5pt}
\renewcommand{\arraystretch}{1.15}
\begin{table}[tb]
\centering
\begin{tabular}{|l c S[table-align-uncertainty, table-format=2.1(3)] S[table-align-uncertainty, table-format=2.1(3)] S[table-align-uncertainty, table-format=2.1(3)] S[table-align-uncertainty, table-format=2.1(3)]|}
\hline 
 & \#OPs&{Accuracy}&{Recall}&{Precision}&{$F_1$ score}\\
\hline 
\multirow{3}{*}{No pretraining}&20&78.8 (125) &72.3(114)&73.4(129)&67.8(141)\\
&40&88.8(77)&83.2(84)&83.8(92)&80.4(103)\\
&60&89.7(66)&82.8(94)&85.8(76)&80.8(103)\\ 
\hline 
\multirow{3}{*}{Contrastive}&20&84.4(106)&77.2(84)&78.8(53)&73.9(89)\\
&40&91.7(55)&85.4(61)&88.2(56)&83.8(71)\\
&60&92.0(45)&86.2(42)&85.5(48)&83.6(49)\\ 
\hline 
\multirow{3}{*}{Ranking}&20&86.1(72)&79.4(65)&82.9(59)&77.2(74)\\
&40&90.2(64)&85.6(62)&85.2(59)&82.5(73)\\
&60&90.3(54)&85.2(62)&86.1(52)&82.9(69)\\ 
\hline
\multirow{3}{*}{\shortstack[l]{1\textsuperscript{st} \& 2\textsuperscript{nd} order \\ contrastive}}&20&88.1(58)&80.7(57)&83.8(56)&78.6(61)\\
&40&90.7(104)&86.3(75)&86.9(61)&83.4(101)\\
&60&92.7(43)&87.0(40)&87.6(53)&84.6(54)\\ 
\hline 
\end{tabular}
\caption{Performance of the baseline (first row) and the pretrained models on the surgical phase segmentation task. \emph{\#OPs} denotes how many labeled OPs were used.}
\label{tab:comparison:methods}
\end{table}

\begin{table}[tb]
\centering
\begin{adjustbox}{center}
\resizebox{\textwidth}{!}{%
\begin{tabular}{|l c S[table-align-uncertainty, table-format=2.1(3)] S[table-align-uncertainty, table-format=2.1(3)] S[table-align-uncertainty, table-format=2.1(3)] S[table-align-uncertainty, table-format=2.1(3)] S[table-align-uncertainty, table-format=2.1(3)] S[table-align-uncertainty, table-format=2.1(3)] S[table-align-uncertainty, table-format=2.1(3)] |}
\hline
&{\#OPs}&{P1}&{P2}&{P3}&{P4}&{P5}&{P6}&{P7}\\
\hline
\multirow{3}{*}{\shortstack[l]{No pre-\\ training}}&20&64.5(357)&83.4(159)&59.0(334)&80.8(142)&62.0(240)&62.8(202)&62.4(183)\\
&40&88.7(216)&92.3(108)&75.3(294)&90.3(143)&74.0(176)&71.7(194)&71.0(143)\\
&60&82.4(253)&94.9(57)&81.1(185)&92.0(109)&76.1(153)&73.7(179)&65.4(246)\\
\hline
\multirow{3}{*}{\shortstack[l]{1\textsuperscript{st} \& 2\textsuperscript{nd}\\order\\contrastive}}&20&79.3(257)&92.2(76)&81.9(144)&91.3(80)&72.4(175)&72.8(175)&60.0(244)\\
&40&87.6(153)&95.7(71)&86.3(133)&91.4(181)&78.5(184)&73.2(217)&71.6(181)\\
&60&90.2(147)&97.6(27)&89.3(99)&95.9(37)&75.4(193)&76.9(182)&67.8(166)\\
\hline
\end{tabular}
}
\end{adjustbox}
\caption{Comparison of the baseline and the best performing pretrained model. We report the average $F_1$ scores calculated for each of the phases P1 to P7.} 
\label{tab:comparison:phases}
\end{table}

\section{Discussion}
Table \ref{tab:comparison:methods} clearly shows that all three pretrained models outperform the baseline when being fine-tuned on the same set of labeled training data. The performance boost is especially apparent when only 20 labeled videos are available. Here, in terms of $F_1$ score, pretraining achieves an increase from 67.8 to up to 78.6 while halving the standard deviation. Pretraining still improves performance when more labeled videos are available. Notably, the pretrained models fine-tuned on only 40 labeled videos outperform the baseline trained on 60 videos. 
We conclude that the proposed SFA-based pretraining enables a CNN to learn feature representations that are beneficial to the task of surgical phase segmentation.

Comparing the three pretraining variants, we do not find big differences.
All in all, using a combination of first and second order temporal coherence for pretraining seems to offer the largest boost to performance, especially when only few (20) labeled videos are used.

Looking at the results with respect to each surgical phase (table~\ref{tab:comparison:phases}), we see that most phases benefit greatly from pretraining 
(variant \textbf{1\textsuperscript{st} \& 2\textsuperscript{nd} order contrastive}) when only 20 labeled videos are available. The effect of pretraining diminishes when the number of labeled videos is increased, but is still noticeable in the majority of phases.
Only the benefit to phase P7 seems negligible.

P7 contains visual similarities with P5 and P6, which makes them difficult to distinguish.
Since the phase is short (about 1 to 3 min), frames that we label as close during pretraining may belong to previous phases.
Likewise, frames that belong to previous phases but are temporally close are not selected as distant pair.
Hence, the network learns features that are rather invariant than discriminative with regard to phase P7 and P6 or P5. 

To shed some light on the features learned during pretraining, we investigated which images the network considers similar.
We selected query frames $\lbrace I^q \rbrace$ from a video used during pretraining. Then, for each frame $I^q$ and each video~$v$ in the test set, we identified the frame $I^{q, v}$ in $v$ that is most similar to $I^q$, i.e., closest to $I^q$ in feature space. More formally, $I^{q, v} = \argmin_{I_t \in v} D(f(I^q), f(I_t))$, where D was chosen to be the L2 norm. To calculate the embedding $f$, we used the \textbf{1\textsuperscript{st} \& 2\textsuperscript{nd} order contrastive} pretrained \textit{FeatureNet} (before fine-tuning).

\begin{figure}[tb]
\begin{center} 
\includegraphics[width=0.95\textwidth]{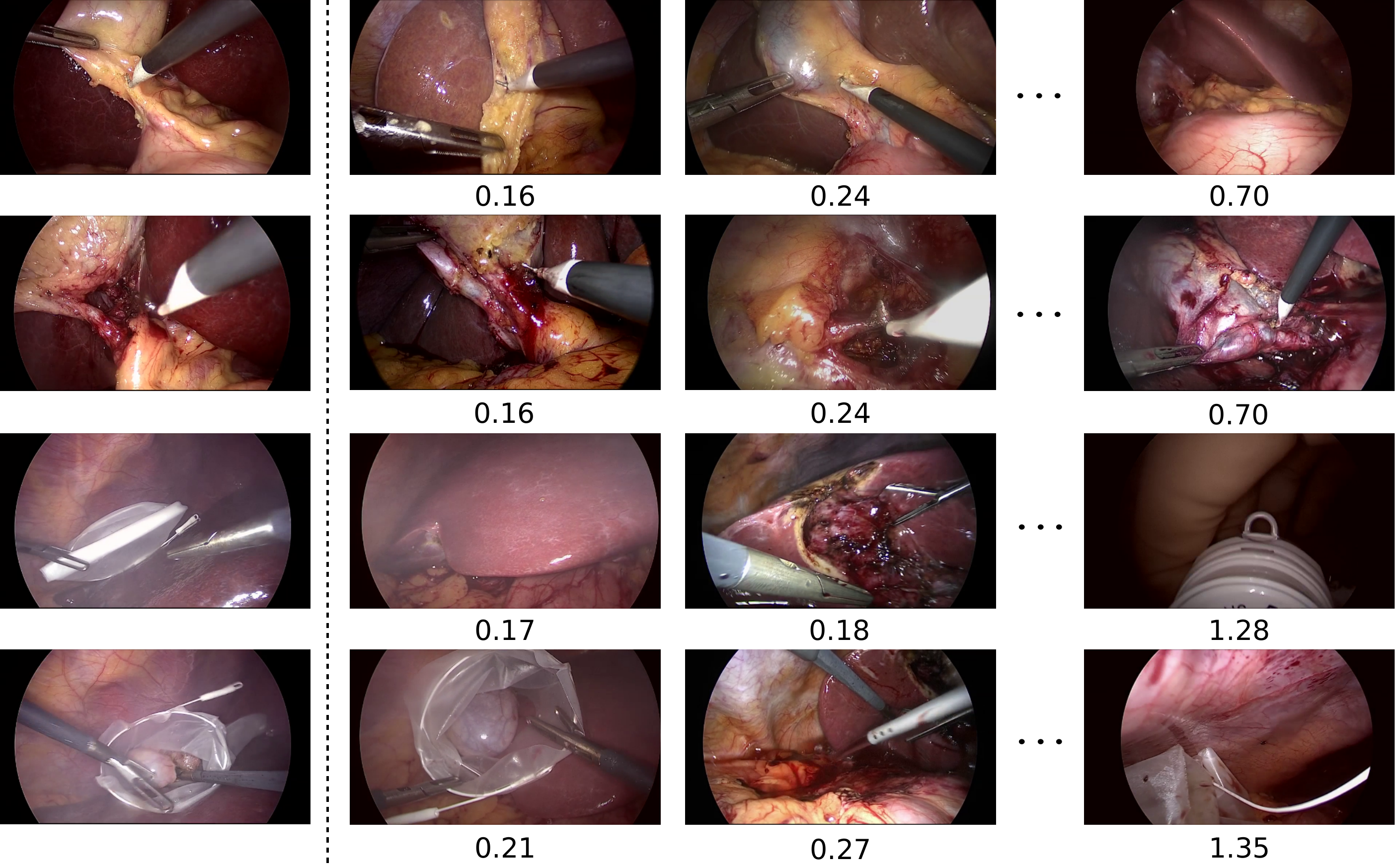}
\caption{Image retrieval task. Each row represents one query. Left-most: Query frame. Right: The frames closest in feature space, one per test video. Numbers denote distance to query frame. The depicted frames are sorted with regard to this distance.}
\end{center}
\label{fig:retrieval}
\end{figure}

Figure \ref{fig:retrieval} presents four selected queries.
Generally, it can be seen that images that are close in feature space show similar scenes with regard to anatomical structures and/or tool presence.
The first and second query frames depict scenes that only differ in the amount of blood visible, a trait also observed in the retrieved frames.
Likewise, the third and fourth query frames show similar scenes. However, the third query frame is unusual as the specimen bag is closed. Observing that the retrieved images are semantically not closely related to the query frame, we assume that its embedding does not reflect the presence of the specimen bag.
For the fourth query frame, which is visually similar but more representative, semantically similar frames are retrieved.

We refrain from comparing temporal coherence-based learning to other pretraining methods for surgical phase segmentation~\cite{Bodenstedt2017,yengera2018less} since these studies were conducted using other datasets, namely EndoVis2015 (7 cholecystectomies) in~\cite{Bodenstedt2017} and 120 cholecystectomies in~\cite{yengera2018less}. 

\section{Summary}
In this paper, we show that the temporal coherence of unlabeled laparoscopic video can be exploited for self-supervised pretraining by training a CNN to map temporally close video frames onto embeddings that are close in feature space.
When extended into a CNN-LSTM architecture for surgical phase segmentation, all pretrained models outperform the non-pretrained baseline when being fine-tuned on the same labeled dataset. 
Using a combination of first and second order temporal coherence, the pretrained models even perform similarly or better than the baseline when less labeled data is used.
Combining our approach with temporal order-based concepts into a more holistic temporal coherence-based pretraining method could possibly enhance the discriminative properties of the learned embedding and improve performance even further.

Future work will address the question whether the learned embeddings can be used for unsupervised detection of more fine-grained video segments, such as surgical activities or steps.
Furthermore, we will investigate whether the notion of slow and steady features is beneficial for regularization during supervised training compared to using the concept during a separate pretraining phase.

%
%
%
%
\bibliographystyle{splncs04}
\bibliography{paper_v2}

\end{document}